\def\year{2019}\relax
\documentclass[letterpaper]{article} 
\usepackage{aaai19}  
\usepackage{times}  
\usepackage{helvet}  
\usepackage{courier}  
\usepackage{url}  
\usepackage{graphicx}  
\usepackage{epsfig}
\usepackage{amsmath}
\usepackage{amssymb}
\usepackage{enumitem}
\usepackage{algorithm}
\usepackage{algpseudocode}
\usepackage{array}
\usepackage{subfigure}
\usepackage{epstopdf}
\usepackage[final]{changes}
\usepackage{hyperref}
\newcommand{\calC}{\mathcal{C}}
\newcommand{\calX}{\mathcal{X}}
\newcommand{\calL}{\mathcal{L}}

\begin{document}
%
\title{Unique Identification of Macaques for Population Monitoring and Control}
\author{Ankita Shukla$^1$\thanks{Equal Contribution}, Gullal Singh Cheema$^1$\footnotemark[1], Saket Anand$^1$, Qamar Qureshi$^2$, Yadvendradev Jhala $^2$ \\
$^1$ IIIT-Delhi, India \\
$^2$ Wildlife Institute of India, Dehradun\\
{\tt\small $^1$\{ankitas, gullal1408, anands\}@iiitd.ac.in, $^2$\{qnq,jhalay\}@wii.gov.in}}

\maketitle
\begin{abstract}
Despite loss of natural habitat due to development and urbanization, certain species like the Rhesus macaque have adapted well to the urban environment. With abundant food and no predators, macaque populations have increased substantially in urban areas, leading to frequent conflicts with humans. Overpopulated areas often witness macaques raiding crops, feeding on bird and snake eggs as well as destruction of nests, thus adversely affecting other species in the ecosystem. In order to mitigate these adverse effects, sterilization has emerged as a humane and effective way of population control of macaques. As sterilization requires physical capture of individuals or groups, their unique identification is integral to such control measures. In this work, we propose the Macaque Face Identification (MFID), an image based, non-invasive tool that relies on macaque facial recognition to identify individuals, and can be used to verify if they are sterilized. Our primary contribution is a robust facial recognition and verification module designed for Rhesus macaques, but extensible to other non-human primate species. We evaluate the performance of MFID on a dataset of 93 monkeys under closed set, open set and verification evaluation protocols. Finally, we also report state of the art results when evaluating our proposed model on endangered primate species.

\end{abstract}

\section{Introduction}\label{sec:intro}
Expansion of urban areas and infrastructural developments has led to depletion of forest cover, which is a natural habitat for many animal species. On one hand, this habitat loss threatens extinction of many species, on the other hand, some resilient ones like the \emph{Rhesus Macaque} have adapted to the urban lifestyle. In urban areas, rhesus macaques have access to abundant food, water and shelter but have no natural predators (like leopards, or other big cats). With a life-span of 25-30 years, and a gestation period of about 165 days \cite{Lang_2005}, their population has grown at an alarming rate in many geographical areas.   

Rhesus macaques are native to countries in South and Southeast Asia, but are also found in many countries in the Americas. Their rapid population growth and notoriously destructive nature has qualified them as an invasive species and an environmental threat \cite{GISD_2018}. In their alien ranges where rhesus macaques were introduced, they have reportedly caused destruction of over 30 acres of mangroves \cite{Florida_anderson2017}, decreased populations of native birds \cite{Birds_2016predation} and contaminated tidal creeks to have elevated levels of fecal coliform bacteria \cite{AndersonHistory_2016}. Native ranges are no exception either, the more aggressive rhesus macaques have spread and become a threat to the bonnet macaque, a species endemic to peninsular India \cite{Kumar2011,Erinjery_2017}. As a consequence, various organizations across countries have taken measures to control their population \cite{AndersonHistory_2016}. 

Beyond environmental damage, Rhesus macaques have had substantial impact on humans, extending from economic losses -- US Dept. of Agriculture (USDA) reported about $1.3$ million dollars annual losses in southwest Puerto Rico associated with crop-raiding by primates \cite{AndersonHistory_2016} -- to health hazards through monkey bites \cite{Imam2013,Florida_anderson2017}, with some countries reporting as many as a 1000 bites a day
\footnote{\href{https://www.usatoday.com/story/news/world/2017/05/11/monkeys-india-pests-contraceptives/101466580/}{USA Today article, May, 11, 2017: Why India is going bananas over birth control for monkeys}}. 
These problems of human-monkey conflicts are exacerbated in densely populated regions where humans and macaques coexist. While recent large-scale, peer-reviewed, studies on the impact of rhesus macaques on humans are limited \cite{Imam2013}, a lot of grey literature and news articles indicate that such conflicts are increasing in magnitude and frequency.  

Given the multi-faceted, adverse impacts of uncontrolled rise in rhesus macaque populations, various remedial steps have been taken. Approaches like relocation \cite{imam_yahya_malik_2002} mitigates the problem locally, the monkey troupes suffer stress, anxiety and trauma while adapting to their new environment \cite{Dettmer_relocate2012}, in-turn making them more aggressive. These commensal monkey troupes, even when relocated to a deep forested area, tend to find the nearest human settlements for their survival \cite{Govindrajan_2015}. On the other hand, culling has been an effective solution in Puerto Rico \cite{AndersonHistory_2016}, but it has social and religious implications in certain cultures. In their anthropocentric opinion survey, \cite{Imam2013} reports about 80\% of the 300 north-Indian participants have religious attachments with monkeys, due to which on occasion, culling has led to strong protests by religious and animal activist groups alike \cite{Govindrajan_2015}. Thus a pest management problem, has become a complex issue due to social and religious constraints. Fortunately, a long-term solution for population control of rhesus macaques has emerged in the form of sterilization. 

A successful example of effective application of sterilization in practice is the Monkey Sterilization Program\footnote{\href{http://hpforest.nic.in/pages/display/ZDRmNjVhiHFmYTU2cw==-monkey-sterilization-programme}{hpforest.nic.in $\rightarrow $ Wildlife $ \rightarrow $ Monkey Sterilization Programme}} implemented by the Himachal Pradesh Forest Department in India. Since its inception in 2006, the Monkey Sterilization Centers (MSC) have cumulatively sterilized over 125,000 rhesus macaque individuals until 2017. As a result of this initiative, the estimated population has reduced from about 317,000 in 2004 to about 226,000 in 2013. The process of sterilization requires capturing the monkey(s) individually or in groups, and transporting them to the MSC, where they undergo surgery, and receive post-op care before being released in their \emph{original} territory. Nearly a quarter of the recurring cost of the sterilization program is spent on catching and transporting the monkeys. Being a tedious task, the local people, administrative bodies and professional monkey catchers work together to identify the capture site, estimate the strength of the monkey troupe, and plan the capture\footnote{The numbers quoted are based on a combination of sources: our research of grey literature, news articles and interactions with forest officials and scientists. The numbers are approximate and the sources are not included for brevity.}.

As the number of sterilized macaques grow, there is an increasing chance of recapturing a sterilized one which leads to wasted resources. Various approaches for visual identification have been used like neck-bands, ear tags and freeze branding, with the latter becoming increasingly popular. In this paper, we propose an alternate, non-invasive approach to identify individual macaques using a handheld camera. An effective visual identification technique could be beneficial at multiple stages of capturing monkeys. The capture site identification as well as the troupe strength estimation can be crowdsourced. Before setting up the capture, an individual could be checked against a database of previously sterilized individuals, thus saving effort and time. 

Inspired by the success of face recognition for humans, in this work, we aim to automate the process of unique identification of macaques. We propose a deep learning based approach for recognizing facial images captured through different imaging devices like cellphone cameras and DSLRs. We summarize the main contributions of this paper here:
\begin{itemize}
	\item Proposed \emph{Macaque Face IDentification} (MFID), an alignment free deep learning system for automated identification of rhesus macaques from images.
	\item Employed an objective function to fine tune a pretrained model for identification of primates using face images.
	\item Demonstrated the generalization of MFID to other species like chimpanzees and achieved state of the art results.
    \item We collected Rhesus Macaques dataset in their natural dwellings and plan to publicly open source the dataset to encourage further research in the field.
\end{itemize}

\section{Previous Work}\label{sec:pw}
The literature for unique individual identification of primates has evolved owing to its relevance to biologists, researchers and conservationists. However, most of the existing literature in the field focuses on endangered species, except for the work by Witham \cite{witham2017automated}, which tackles unique identification of rhesus macaques in videos for behaviour monitoring. The dataset comprises images of 34 adult individuals, captured in an indoor lab setting. Witham's pipeline for face recognition used a Viola-Jones detector \cite{Viola_2004} for detecting faces, an eye and nose detector to further reduce false alarms as well as to register faces, follwed by the SVM and LDA based recognition module. While the performance was good, the approach could handle limited pose variations. This approach is susceptible to alignment errors, particularly when there are failures in the eye or nose detection module. Moreover, the approach is not tested in uncontrolled, outdoor situations. 
 

To the best of our knowledge, all other non-invasive approaches deal with endangered primate species like lemurs and golden monkey, which while relevant, do not report performance on open-range rhesus macaques. We broadly categorize these approaches into two categories: Non Deep Learning Approaches and Deep Learning Approaches. 

\subsection{Non Deep Learning Approaches}
Traditional face recognition pipelines comprised of face alignment, followed by low level feature extraction and classification. Early works in primate face recognition \cite{Loos2012}, adapted the Randomfaces \cite{Robust_2009} technique for identifying chimpanzees in the wild and follows the standard pipeline for face recognition. Later, LemurID was proposed in \cite{LemurID_2017}, which additionally used manual marking of the eyes for face alignment. Patch-wise multi-scale Local Binary Pattern (LBP) features were extracted from aligned faces and used with LDA to construct a representation, which was then used with an appropriate similarlity metric for identifying individuals. 

\subsection{Deep Learning Approaches}
Freytag et al. use Convolutional Neural Networks (CNNs) for learning a feature representation of chimpanzee faces. For increased discriminative power, the architecture uses a bilinear pooling layer after the fully connected layers, followed by a matrix log operation. These features are then used to train an SVM classifier for identification. Later, \cite{Brust_2017} developed face recognition for gorilla images captured in the wild. This approach fine-tuned a YOLO detector \cite{YOLO_CVPR2016} for gorilla faces. For identification, a similar approach was taken as \cite{freytag2016chimpanzee}, where pre-trained CNN features are used to train a linear SVM. More recently, \cite{Jain_2018} proposed an approach for face recognition of multiple primates including lemur, chimpanzees and golden monkey and have shown to achieve state of the art results. However, it requires substantial manual effort to designing landmark templates for face alignment prior to identification process. 
\subsection{Invasive Approaches}
As opposed to the aforementioned non-invasive visual identification techniques, there exist approaches where individuals were identified by attaching a tracking device. For the sake of completeness, we include some of these techniques here. RFIDs \cite{rfid_addali2014} have been used widely for monitoring different species. Similarly, collars \cite{collars_ballesta_2014} and jackets \cite{jacket_Rose2012} have been used for monkeys as well as other species. With advances in visual identification, however, invasive approaches have limitations as they require more effort due to direct involvement of the animals for tagging, maintenance and for incorporating new individuals.
\section{Proposed System}\label{sec:proposed}
We now present our proposed MFID system for unique identification of macaques using facial images. It has two main parts: the identification module, and the detection module. While state of the art deep learning based detectors work sufficiently well, identifying individuals is more challenging. Our contributions are mainly in the design of the loss function of the identification module, motivated by the constraints of the operating environment and ease of use for our end application. We note that the end users of MFID will largely be the general public, professional monkey catchers and field biologists. Typically, we expect the images to be captured in uncontrolled outdoor scenarios, leading to significant variations in facial pose and lighting. These conditions are challenging for robust eye and nose detection, which need to be accurate in order to be useful for facial alignment. Consequently, we train our identification model to work without facial alignment. We acknowledge that the accuracy of face recognition systems improve by alignment, the interaction of multiple learned modules makes debugging and failure analysis more complex. Therefore, we make a design choice by using an identification module that works in an alignment-free setting, and hope to make up for the performance loss by averaging identification scores over multiple probe images of the same target monkey. 

\subsection{Individual Recognition}
One of the primary challenges in training a robust face recognition system for macaques is the lack of large training data. In order to avoid the risk of overfitting, we use an additional pairwise KL-divergence loss term \cite{Kira2015}, which biases the resulting softmax probabilities to be more discriminative for different classes and be similar for the same class. Additionally, by sampling pairs of training samples (both similar and dissimilar), the pairwise term uses the training set more effectively in the qualitative as well as quantitative sense. The proposed loss utilizes pairwise constraints that are generated in the following way.
\subsubsection{Pairwise Constraints}
We use the labeled training data to create similar and dissimilar image pairs. Two images having the same identity label are considered as a similar pair while a dissimilar pair comprises two images with different labels. The set of similar pairs $\calC_s$ and dissimilar pairs $\calC_d$ is given by 
\begin{align}
\nonumber &\calC_s =\{(i, j): x_i, x_j \in \calX, l_i = l_j\},\\
\nonumber &\calC_d =\{(i, j):  x_i, x_j \in \calX, l_i \neq l_j\},\\
&\qquad ~i,~j \in \{1,2,\cdots,n\}
\end{align}
Here, $ \calX $ is the training dataset of $ n $ samples with $ l_i $ as the associated labels. The KL divergence between two distribution $p$ and $q$ corresponding to points $x_p$ and $x_q$ is given by
\begin{align}
\label{eq:KL}
KL(p||q) =\sum_{i=1}^{K}{p_i \log\frac{p_i}{q_i}}
\end{align}
Therefore, for a similar pair, we aim to minimize the symmetric variant of (\ref{eq:KL}) given by
\begin{align}\label{eq: sim}
\calL_{s} =KL(p||q)+KL(q||p)
\end{align}
For the dissimilar pairs, we make use of a large margin style loss to increase the discriminative power of the softmax probabilities. As in (\ref{eq: sim}), the corresponding symmetric loss term for dissimilar pairs is given by 
\begin{align}\label{eq: dsim}
\calL_{d}= \max(0,m-KL(p||q)) + \max(0,m-KL(q||p))
\end{align}
The loss $ \calL_d $ penalizes the model when the KL-divergence of a dissimilar pair is separated by a margin smaller than $ m $. 
\subsubsection{Loss Function and Optimization}
We augment the standard cross entropy loss $ \calL_{ce} $, with the pairwise loss terms defined over similar and dissimilar pairs to enforce similar class distribution for images of same class and dissimilar distributions for images of different class. 
\begin{align}
\calL(\theta) = \calL_{ce} + \frac{1}{|\calC_s|} \sum_{j,k \in \calC_s}{\calL_s} + \frac{1}{|\calC_d|} \sum_{j,k \in \calC_d}{\calL_d}
\end{align}


\subsection{Detection}
We use state-of-the-art Faster R-CNN \cite{Ren_NIPS_FRCNN2015} object detector that has performed really well on several benchmark datasets for detection and objection recognition. The detector is composed of two modules in a single unified network. The first module is a deep CNN that works as a Region Proposal Network (RPN) and proposes regions of interest (ROI), while the second module is a Fast R-CNN [10] detector that categorizes each of the proposed ROIs. We initialize the network with pre-trained ImageNet weights and fine-tune it for detection of monkey faces.
\section{Experimental Results}\label{sec:exp}
In this section, we present the experimental evaluation of the MFID system. The following sections will describe the datasets used, different evaluation protocols employed, the experimental results and analysis. In addition to macaque faces, we also report results on a publicly available chimpanzee face dataset and show comparative results with state of the art.

\begin{figure*}[h!]
	\centering
			\subfigure[C-Zoo]{\includegraphics[width =0.235\textwidth,scale=1]{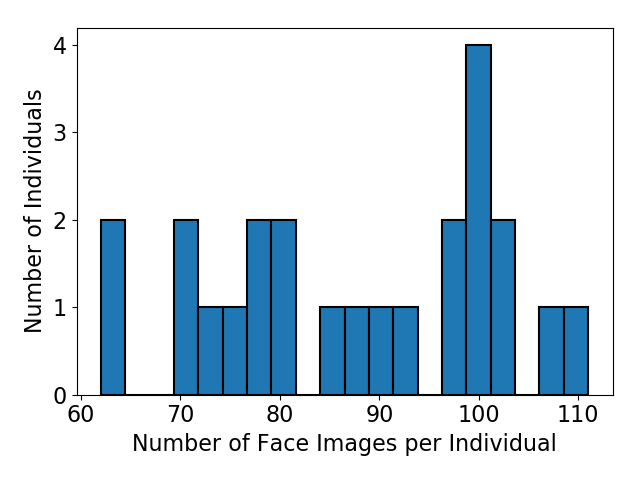}  }
        \subfigure[C-Tai]{\includegraphics[width  =0.235\textwidth,scale=1]{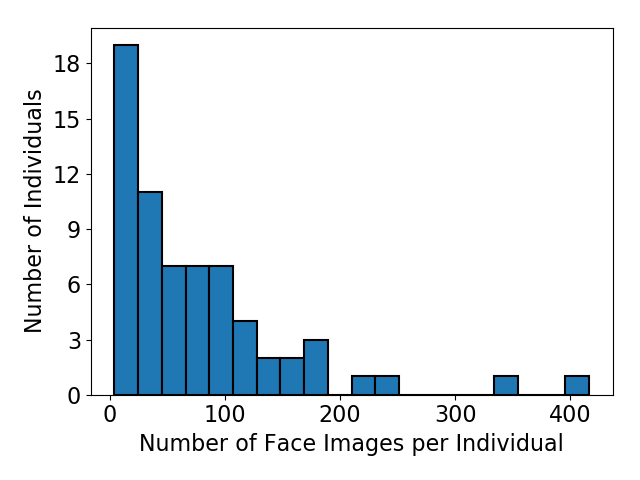} }
        \subfigure[C-Zoo+C-Tai]{ \includegraphics[width =0.235\textwidth,scale=1]{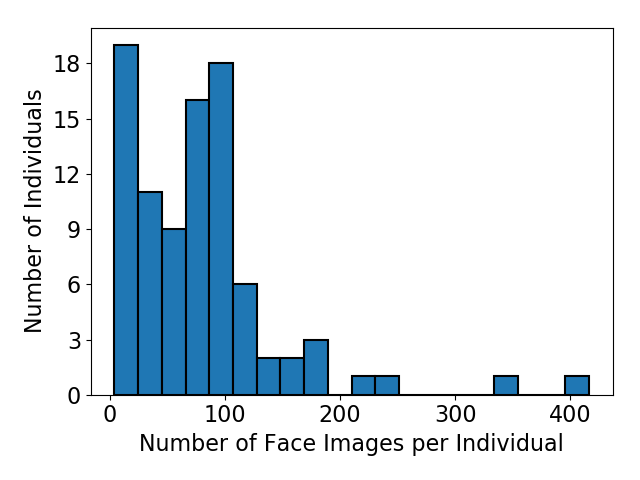} }
        \subfigure[Rhesus Macaques]{ \includegraphics[width  =0.235\textwidth,scale=1]{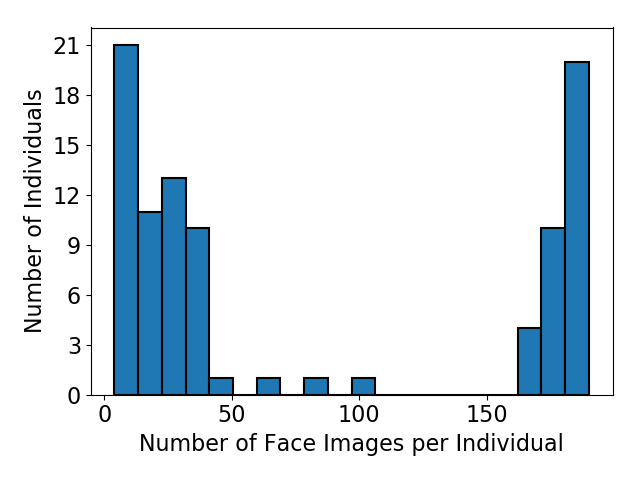}
        }
		\caption{Histogram of the number of face images per identity in (a) C-Zoo, (b) C-Tai, (c) C-Zoo+C-Tai and (d) Rhesus Macaques datasets. The total number of identities in CZoo, C-Tai, CZoo+CTai and Rhesus Macaque is 24, 66, 90 and 93 respectively. }
		\label{fig: data_dist}
\end{figure*}

\subsection{Dataset Description}
We evaluate our model using three different primate species data, the details of which are given in Table \ref{tab: datasets}. As is typical of wildlife data collected in uncontrolled environments, all the three datasets have a significant class imbalance, which is shown using the histogram plots in figure \ref{fig: data_dist}. 
\subsubsection{Rhesus Macaques Dataset:} 
\begin{figure*}[h]
	\centering
			{\includegraphics[width =0.2\textwidth,height=2cm,keepaspectratio,scale=1]{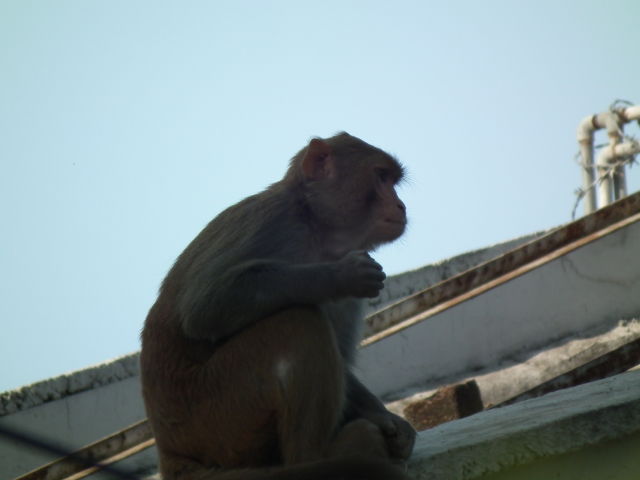}
        \includegraphics[width  =0.2\textwidth,height=2cm,keepaspectratio,]{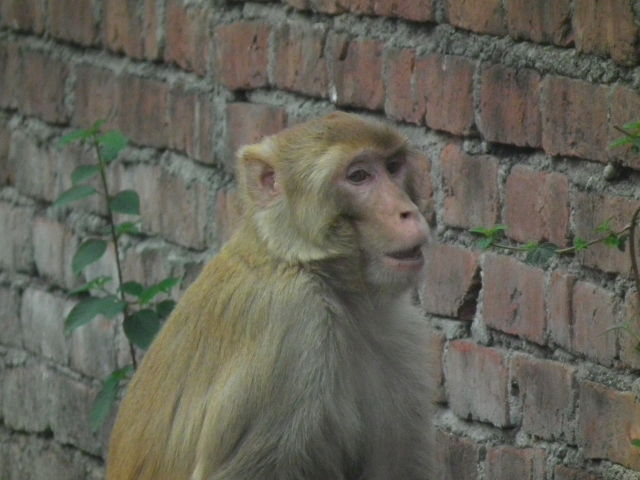}
       \includegraphics[width =0.2\textwidth,height=2cm,keepaspectratio]{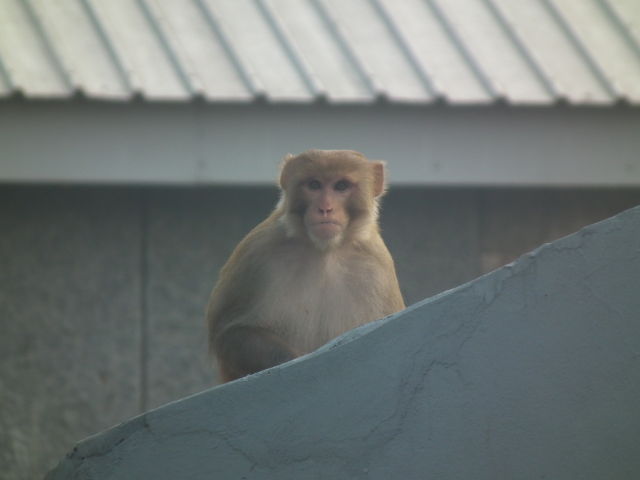}
       \includegraphics[width  =0.2\textwidth,height=2cm,keepaspectratio]{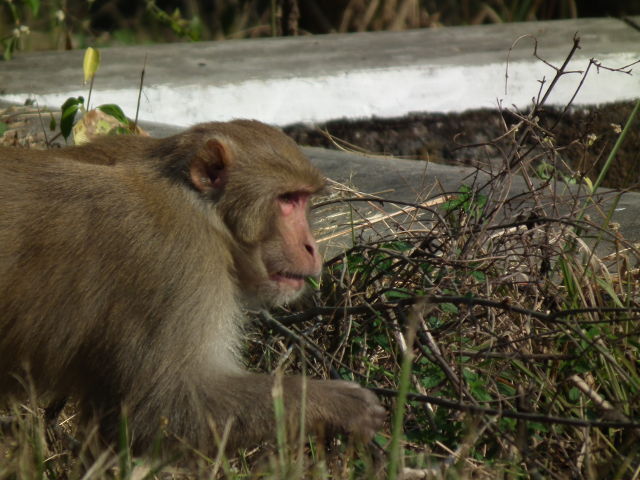}
        \includegraphics[width  =0.2\textwidth,height=2cm,keepaspectratio]{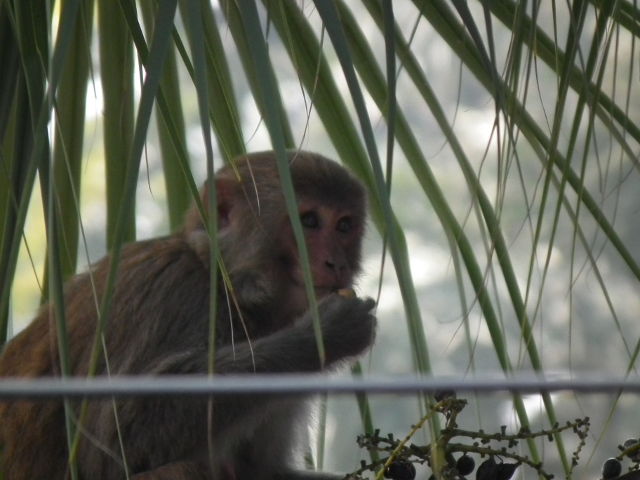}\\
          \includegraphics[width  =0.2\textwidth,height=2cm,keepaspectratio]{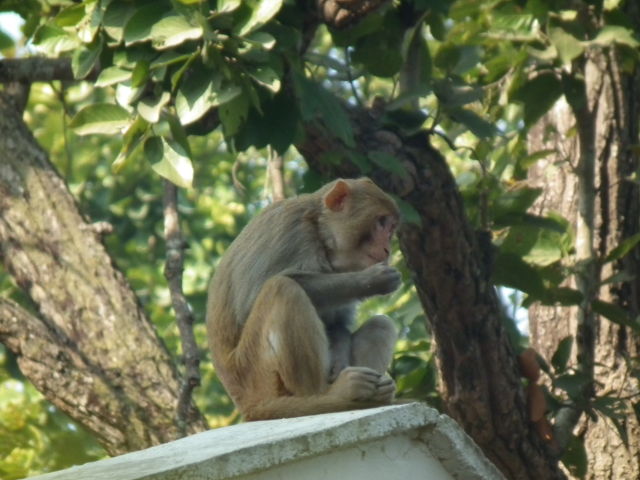}
           \includegraphics[width  =0.2\textwidth,height=2cm,keepaspectratio]{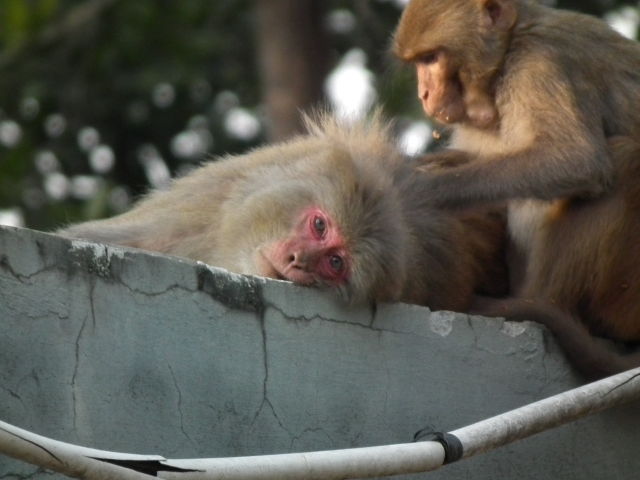}
            \includegraphics[width  =0.2\textwidth,height=2cm,keepaspectratio]{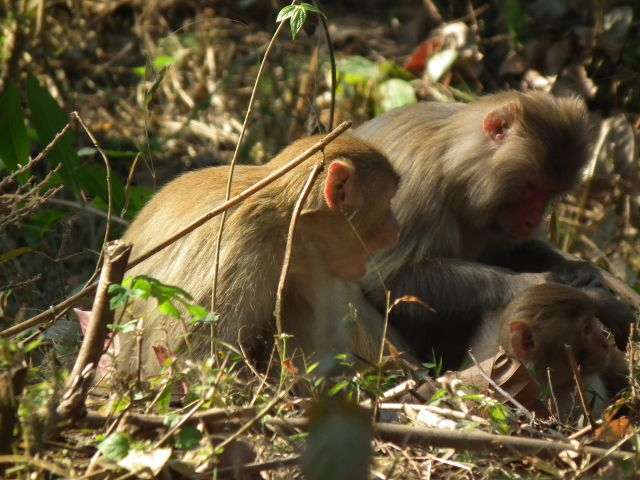}
             \includegraphics[width  =0.2\textwidth,height=2cm,keepaspectratio]{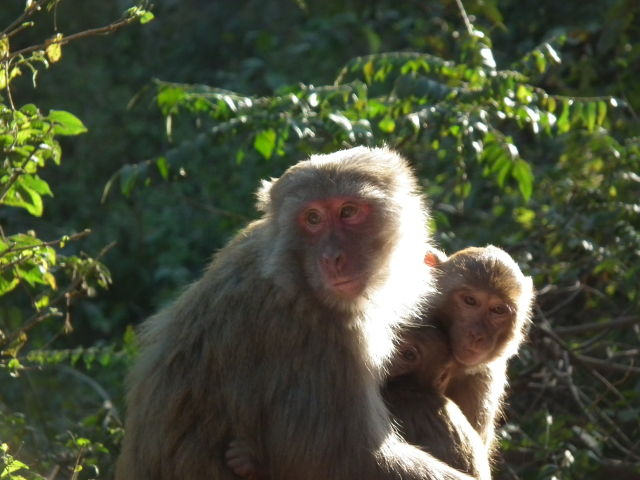}
            \includegraphics[width  =0.2\textwidth,height=2cm,keepaspectratio]{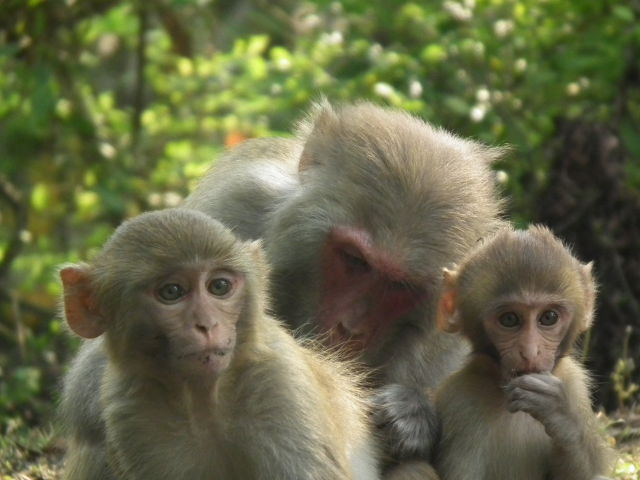}
        }
		\caption{{Example Images of Rhesus Macaques from our dataset collected in Uttarakhand state of India}}
		\label{fig: monk_full_example}
\end{figure*}

\begin{figure*}[h]
	\centering
			{\includegraphics[width =0.2\textwidth,height=1cm,keepaspectratio,scale=1]{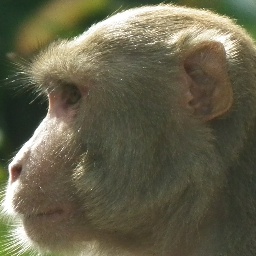}
        \includegraphics[width  =0.2\textwidth,height=1cm,keepaspectratio,]{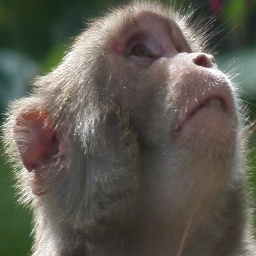}
       \includegraphics[width =0.2\textwidth,height=1cm,keepaspectratio]{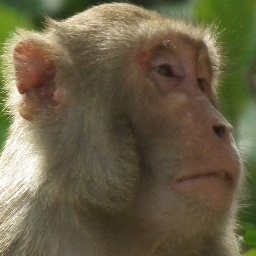}
       \includegraphics[width  =0.2\textwidth,height=1cm,keepaspectratio]{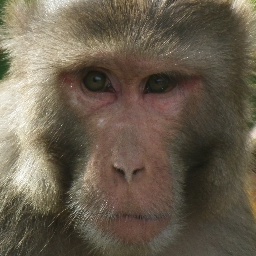}
        \includegraphics[width  =0.2\textwidth,height=1cm,keepaspectratio]{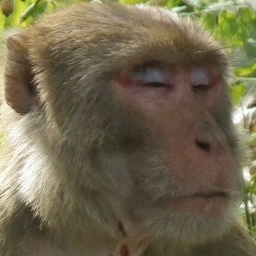}
        }
        {\includegraphics[width =0.2\textwidth,height=1cm,keepaspectratio,scale=1]{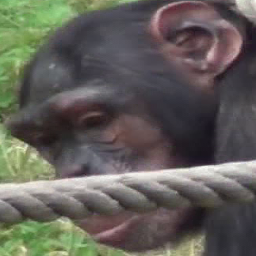}
        \includegraphics[width  =0.2\textwidth,height=1cm,keepaspectratio,]{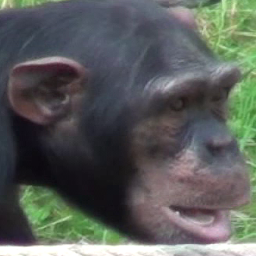}
       \includegraphics[width =0.2\textwidth,height=1cm,keepaspectratio]{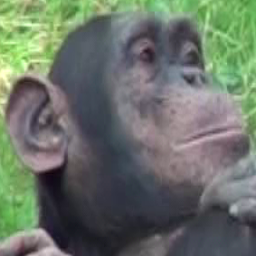}
       \includegraphics[width  =0.2\textwidth,height=1cm,keepaspectratio]{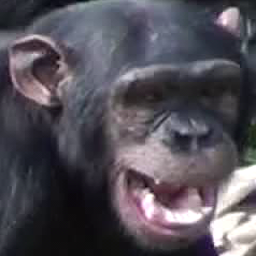}
        \includegraphics[width  =0.2\textwidth,height=1cm,keepaspectratio]{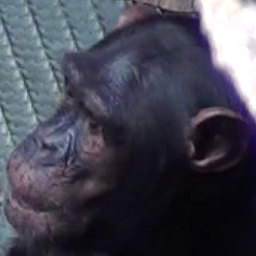}
        }
        
		\caption{{Pose variations for one of the Rhesus Macaque (Left) and Chimpanzee (Right) from the dataset}}
		\label{fig: monk_example}
\end{figure*}

Macaque dataset is collected using DSLRs in their natural dwelling in an urban region in the state of Uttarakhand in northern India. The dataset is cleaned manually to remove images with no or very little facial content (e.g., extreme poses with only one ear or only back of head visible). The filtered dataset had 59 identities with a total of 1399 images. A few sample images from this dataset are shown in figure \ref{fig: monk_full_example}, and an illustrative set of pose variations are shown using the cropped images in figure \ref{fig: monk_example}. Due to the small size of this dataset, we combined our dataset with the publicly available dataset by Witham \cite{witham2017automated}. The combined dataset consists of 93 individuals with a total of 7679 images. Note that we use the combined dataset only for the individual identification experiments, as the public data by Witham comprises of pre-cropped images. On the other hand, the detection and the complete MFID pipeline is evaluated on a test set comprising full images from our macaque dataset.  
\subsubsection{Chimpanzee Dataset}
C-Zoo and C-Tai dataset consists of 24 and 66 individuals with 2109 and 5057 images respectively \cite{freytag2016chimpanzee}. The C-Zoo dataset contains good quality images of chimpanzees taken in a Zoo, while the C-Tai dataset contains more challenging images taken under uncontrolled settings of a national park. We combine these two datasets to get 90 identities with a total of 7166 images.
\begin{table}[h!]
{\small 
	\centering
		\begin{tabular}{|cccc|}
		\hline
  Dataset &  Rhesus Macaques &  C-Zoo& C-Tai  \\ \hline
 \# Samples &7679 &2109 &  5057  \\
 \# Classes & 93 & 24&66  \\
 \# Samples/individual & [4,90]& [62,111]& [4,416]\\\hline
        	\end{tabular}
	\caption{Dataset Description}
	\label{tab: datasets}
    }
 \end{table}
\begin{table*} [h!]
\centering
\begin{tabular}{|c|c|c|c|c|} 
   \hline
   \textbf{Network} & \textbf{CZoo} &\textbf{CTai} & \textbf{CZoo+CTai} & \textbf{Rhesus Macaque} \\
   \hline
   \textbf{Resnet-18} &  76.12 $\pm$ 2.11  & 53.54 $\pm$ 1.9 &58.74 $\pm$ 1.13  & 87.02 $\pm$ 0.5  \\ 
   \hline
   \textbf{Resnet-50} & 81.58 $\pm$ 1.58 & 57.7 $\pm$ 1.43 &64.96 $\pm$ 1.08& 88.68 $\pm$ 0.72\\
   \hline
   \textbf{Alexnet FC2} & 79.56 $\pm$ 1.65 & 61.68 $\pm$ 1.93 & 65.96 $\pm$ 1.22 &  87.26 $\pm$ 0.84\\ 
   \hline
   \textbf{Alexnet Pool5} & \textbf{87.58 $\pm$ 1.82} &  \textbf{68.16 $\pm$ 2.43} & \textbf{73.46 $\pm$ 0.99} & \textbf{96.04 $\pm$ 0.82} \\
   \hline 
   \textbf{Densenet-121} & 82.04 $\pm$ 1.5 &  57.52 $\pm$ 1.16 & 63.48 $\pm$ 1.29 &  88.04 $\pm$ 0.79\\
   \hline
\end{tabular}
\caption{Classification Performance of CNN+PCA+SVM on CZoo, CTai, CZoo+CTai and Rhesus Macaques}
\label{tab: cnn_svm_1}
\end{table*}
\begin{table*}
	\centering
		\begin{tabular}{|ccccc|}\hline
        Method & Classification& Closed-set & Open-set & Verification \\
        &Rank-1 & Rank-1& Rank-1& 1 \%  FAR \\\hline 
       Baseline (Alexnet Pool5) & 96.04 $\pm$ 0.82 & 91.75 $\pm$ 2.62 & 93.17 $\pm$ 0.93 & 75.59 $\pm$ 5.91 \\
       \hline
       ResNet-18+Cross Entropy &97.97 $\pm$0.70 & 96.52$\pm$ 1.95& 98.08 $\pm$ 0.45 & 92.72 $\pm$2.45 \\\hline
       ResNe-18+MFID &98.73$\pm$0.38 &97.06 $\pm$ 2.4 & 97.42 $\pm$ 1.02 & 96.82 $\pm$1.63\\\hline
       DenseNet-121+Cross Entropy & 98.03 $\pm$ 0.49& 95.8 $\pm$ 2.18& 98.3 $\pm$ 0.75 &94.64$\pm$3.68 \\\hline
       DenseNet-121+MFID &\textbf{98.83$\pm$0.45} &\textbf{97.19$\pm$ 2.64}& \textbf{98.98 $\pm$ 0.37} & \textbf{97.5 $\pm$1.54}\\\hline
       \end{tabular}
        \caption{Evaluation of Rhesus Macaque dataset for  classification, closed set, open set and verification setting; Baseline results are CNN+PCA+SVM achieved by the best network}
        \label{tab:maca_results}
\end{table*}
\subsection{Network Parameters and Training}
We resize all cropped macaque face images to $224 \times 224$. For C-Zoo and C-Tai, the images are resized to $256 \times 256$. We add the following data augmentations: random horizontal flips and random rotations within $5$ degrees for both the datasets in addition to random crops of $224 \times 224$ only for the Chimpanzee data. We use the following base network architectures for MFID: ResNet \cite{ResNet18_2016} with 18 and 50 layers and DenseNet \cite{DenseNet} with 121 layers. We also use AlexNet \cite{Krizhevsky_2012} for generating baseline results. For fine-tuning the different networks with cross entropy loss we used a batch size of 32, and for MFID that uses pairs of samples, a batch size of 16 pairs is used that results in 32 images. We used SGD for optimization with an initial learning rate of $10^{-3}$. The number of epochs used for training for Macaques and C-Tai dataset is 50. For combined dataset (CZoo+CTai), the number of epochs are 60. In each of the cases, we reduce the learning rate every 20 epochs by $0.1$. Since, the dataset of C-Zoo is small, we fine-tuned the networks for 30 epochs while reducing the learning rate every 10 epochs.
\subsection{Evaluation Protocol}
We evaluate and compare the performance of our MFID system under four different experimental settings, namely: classification, closed set, open set and verification.
\subsubsection{Classification}
To evaluate the classification performance the dataset is divided into $80\%/20\%$ train/test splits. We present the mean and standard deviation of classification accuracy over five stratified splits of the data. As opposed to other evaluation protocols discussed below, all the identities are seen during the training, with unseen samples of same identities in the test set.
\subsubsection{Open and Closed Set Identification}
Both, closed set and open set performance is reported on unseen identities. We perform 80/20 split of data w.r.t. to identities, which leads to a test set with 18 identities in test for both chimpanzee and macaque datasets. We again use five stratified splits of the data. For each split, we further perform 100 random trials for generating the probe and gallery sets. However, the composition of the probe and gallery sets for the closed set scenario is different from that of open set. \\
\noindent\textbf{Closed Set}: In case of closed set identification, all identities of images present in the probe set are also present in the gallery set. Each probe image is assigned the identity that yields the maximum similarity score over the entire gallery set. We report the fraction of correctly identified individuals at Rank-1 to evaluate the performance of closed set. \\
\noindent\textbf{Open Set}: In case of open set identification, some of the identities in the probe set may not be present in the gallery set. This allows to evaluate the recognition system to validate the presence or absence of an identity in the gallery. To validate the performance, from the test of 18 identities, we used all the images of 6 identities as probe images with no images in the gallery. The rest of the identities are partitioned in the same way  as closed set identification to create probe and gallery sets. We report Detection and Identification Rate (DIR) at $1\%$ FAR to evaluate open set performance.  
\subsubsection{Verification}
We compute positive and negative scores for each sample in test set. The positive score is the maximum similarity score of the same class and negative scores are the maximum scores from each of the classes except the class of the sample. In our case, where the test data has 18 identities, each sample is associated with a set of 18 scores, with one positive score from the same identity and 17 negative scores corresponding to remaining 17 identities. The verification accuracy is reported as mean and standard deviation at $1\%$ False Acceptance Rates (FARs).

\begin{figure*}[h!]
	\centering
        \subfigure[C-Zoo+C-Tai: (Left) CMC, (Right) TAR vs FAR]{ \includegraphics[width =0.24\textwidth,scale=1]{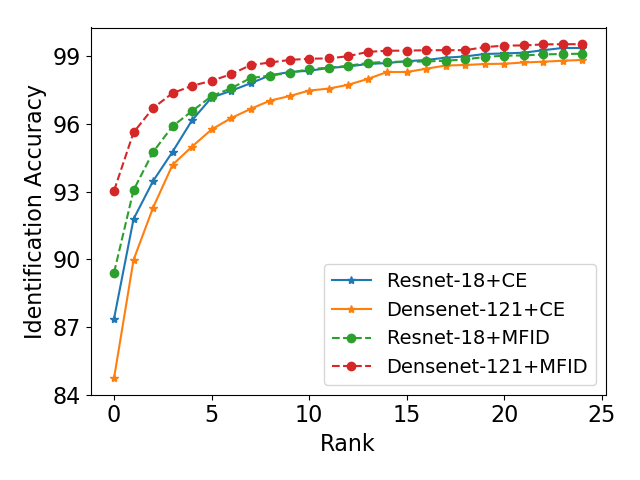} \includegraphics[width =0.24\textwidth,scale=1]{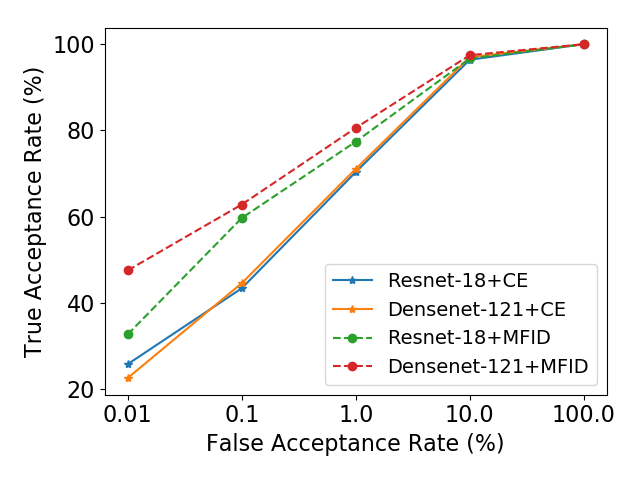} }
         \subfigure[Rhesus Macaques: (Left) CMC, (Right) TAR vs FAR]{ \includegraphics[width =0.24\textwidth,scale=1]{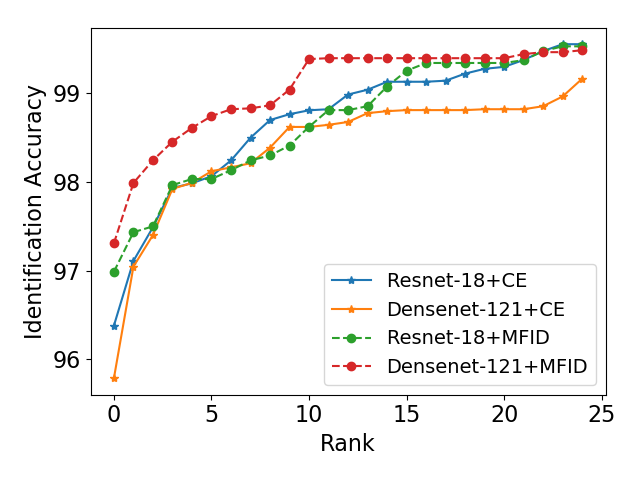} \includegraphics[width =0.24\textwidth,scale=1]{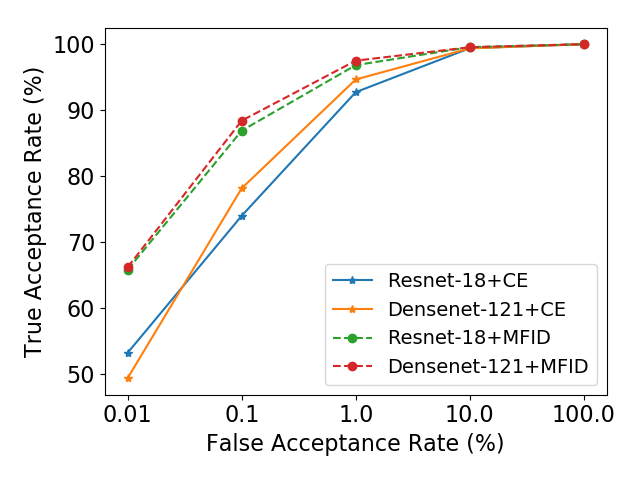}}
       \vspace{-0.1cm}	
		\caption{CMC and TAR vs FAR plots for (a) C-Zoo+CTai and (b) Rhesus Macaques dataset}
		\label{fig: cmc_tar}    
\end{figure*}
\begin{table*}[h!]
	\centering
		\begin{tabular}{|c|c|c|c|c|}
		\hline
        Method & Classification & Closed-set & Open-set & Verification \\
         & Rank-1 & Rank-1& Rank-1& 1 \%  FAR \\\hline 
         Baseline (Alexnet Pool5) & 73.46 $\pm$ 0.99 & 80.57 $\pm$ 3.32 & 76.43 $\pm$ 5.37 & 63.16 $\pm$ 2.42 \\
         \hline
        SphereFace-20 \cite{Liu_Sphereface2017} & N/A & 75.49$\pm$3.80 & 30.75$\pm$12.41 & 48.62$\pm$6.23\\\hline
        SphereFace-4 \cite{Liu_Sphereface2017} & N/A & 74.19$\pm$3.74 & 35.85$\pm$8.22 & 53.92$\pm$2.57\\\hline
        FaceNet \cite{Schroff_FaceNet2015} & N/A & 59.75$\pm$8.64&4.86$\pm$3.38 &17.89$\pm$7.93 \\
        \hline
        PrimNet\cite{Jain_2018} & N/A & 75.82$\pm$1.25 & 37.08$\pm$11.22 & 59.87$\pm$3.34 \\ 
        \hline
       ResNet-18 + Cross Entropy & 83.21 $\pm$ 1.03 & 86.95 $\pm$3.44 & 81.85 $\pm$ 5.52 & 70.36$\pm$ 7.6 \\
       \hline
       DenseNet-121 + Cross Entropy & 84.29 $\pm$ 1.05  & 85.24$\pm$7.30 & 81.15 $\pm$ 6.66 & 71.09$\pm$10.85 \\
       \hline
       ResNet-18 + MFID & 88.85 $\pm$ 0.55 & 89.66$\pm$ 4.35 & 85.85 $\pm$ 5.19 & 77.41 $\pm$ 6.47 \\
       \hline
       DenseNet-121 + MFID & \textbf{90.27 $\pm$ 0.37}  & \textbf{90.38 $\pm$ 5.43} & \textbf{87.63 $\pm$ 6.04} & \textbf{80.59 $\pm$ 8.04}\\
       \hline
       \end{tabular}
        \caption{Evaluation of chimpanzee dataset (C-Zoo +C-Tai) for  classification, closed set, open set and verification setting. Baseline results are CNN+PCA+SVM achieved by the best network. For all other methods, we report the quoted results from (Deb et al. 2018). We follow the same evaluation protocol, except for the open set.}
        \label{tab: chimp_results}
\end{table*}
\begin{table*}[h!]
	\centering
		\begin{tabular}{|c|c|c|c|}
		\hline
        Method & Closed-set & Open-set & Verification \\
        & Rank-1& Rank-1& 1 \%  FAR \\\hline 
       ResNet-18 + Cross Entropy & 97.50 & 95.10 & 85.33 \\\hline
       ResNe-18 + MFID & 97.60 & 95.80 & 89.78 \\\hline
       DenseNet-121 + Cross Entropy & 97.30 & 96.00  & 93.33  \\\hline
       DenseNet-121 + MFID & \textbf{97.70} & \textbf{97.20} & \textbf{96.00} \\\hline
       \end{tabular}
        \caption{Evaluation of Detected Macaque faces for closed set, open set and verification setting}
        \label{tab: detected_maca_results}
\end{table*}
\begin{table*}
	\centering
		\begin{tabular}{|c|c|c||c|c||c|c||}
		\hline
       & \multicolumn{2}{c|}{C-Zoo $\rightarrow$ C-Tai}& \multicolumn{2}{c|}{C-Tai $\rightarrow$ Zoo}& \multicolumn{2}{c|}{C-Zoo+C-Tai $\rightarrow$ Rhesus Macaques}\\\cline{2-7}
       & Cross.Ent & MFID & Cross.Ent& MFID & Cross.Ent & MFID \\\hline
       Closed Set & 54.69 & \textbf{65.92} & 82.48  & \textbf{90.75} & 81.03 & \textbf{86.94}  \\
       \hline
       Open Set & 54.23 & \textbf{64.93} & 83.39 & \textbf{89.39} & 79.17 & \textbf{82.61} \\
       \hline
       Verification & 48.74 & \textbf{57.74} & 64.58 & \textbf{78.38} & 66.81 & \textbf{70.73}  \\
       \hline
\end{tabular}
\caption{Evaluation of learned model across datasets. Left of the arrow indicates the dataset on which the model was trained on, and right of the arrow indicates the evaluation dataset. All the results are reported for DenseNet-121 network}
\label{tab:transfer}
 \end{table*}
        
\subsection{Baseline Results}
For all the baseline experiments, we apply a PCA (Principal Component Analysis) based dimensionality reduction and use principal components that explain $99\%$ of the energy.
We also $L2$-normalize the final features and use a logistic regression classifier with $L2$ regularization. Value of regularization parameter C is varied in the range [1e-5, 1e+5]. The results for different pre-trained CNNs and different layers can be seen in Table \ref{tab: cnn_svm_1}.
All the results are averaged over 5-fold stratified splits. We keep the same parameters as used in the paper \cite{freytag2016chimpanzee}.
\subsection{Results on Macaque dataset}
The results for Rhesus Macaques for all the four evaluation protocol are given in Table \ref{tab:maca_results}. Baseline results reported in the table are the best across the different models. In case of Macaques, AlexNet Pool5 features with PCA with 95 \% energy achieved the best results. We also compare our MFID objective function with Cross Entropy loss. We observe an increase in performance for the four evaluation protocol with a MFID loss as oppose to traditional cross entropy fine-tuned network. Imposing  a KL-divergence loss has improved the discriminativeness of features by skewing the probability distributions of similar and dissimilar pairs. The corresponding CMC and TAR vs FAR plots are shown in Figure \ref{fig: cmc_tar}b.
\subsection{Results on Chimpanzee Dataset}
While, the focus of this work is towards identification of rhesus macaques, we also show the effectiveness of proposed system on chimpanzees and compared with existing approaches in Table 
\ref{tab: chimp_results}. We reported the results of these approaches from PrimNet paper \cite{Jain_2018}. Here SphereFace and FaceNet are human face recognition systems that are fine-tuned on chimpanzee dataset. The results show that the proposed MFID objective function outperforms state of the art results present in the literature. The corresponding CMC and TAR vs FAR plots are shown in Figure \ref{fig: cmc_tar}a.
\subsubsection{Feature learning and Generalization}
To further show the effectiveness of MFID loss function and robustness of features, we perform cross dataset experiments in Table \ref{tab:transfer}. We used model trained on one chimpanzee dataset and extracted features on other chimpanzee dataset to evaluate the performance for closed set, open set and verification task. We compared the quality of the features with the features learned with cross entropy based fine tuning. The results clearly highlight the advantage of MFID over Cross entropy loss for across data generalization. Further, we also extended the results with model trained on chimpanzee dataset and evaluated for Rhesus macaques.
\subsection{MFID System Evaluation}
The above results evaluated the individual performance of recognition stage. The results for identification in Table \ref{tab:maca_results} are obtained with true bounding box of the test samples. As we only have full images of Rhesus Macaques obtained from Uttarakhand (India), we evaluate the detector as well as test the MFID identification stage on these images only from one of the splits. For training the detector, we have total 1191 images, and split it into 80/20 for training/testing. We report the detection performance of Faster R-CNN for monkey faces with 3 metrics, first, MAP (Mean Average Precision) which is a standard metric for evaluating the overlap between predicted boxes and ground truth boxes, second, True positive rate (TPR) and third, False positive rate (FPR). On the test set, we obtain an MAP of 0.952, TPR of 99.56\% and FPR of 0.8\%. The identification stage evaluation using the cropped faces obtained from the detector is shown in Table \ref{tab: detected_maca_results}. For identification evaluation, we have 10 identities and 227 images for both closed-set and verification, whereas for open-set we extend the probe set by adding 8 identities and 1100 samples which are not part of the Uttarakhand dataset.
\section{Conclusion and Future Work}\label{sec:conc}
Our work in this paper is inspired by the impact of overpopulation of Rhesus macaques, which adversely affects the ecosystem as well as puts humans at health and economic risk. Various national and global organizations are taking measures to control their population. While culling of rhesus macaques has been a choice in certain geographical regions, other locations and cultures impose challenges on implementing such measures. Sterilization has emerged as an alternate strategy that controls the macaque population, and will help humans co-exist with macaques in the long-term. Existing techniques of sterilization that are scalable and practical are invasive, requiring the capture of the animals. Capturing macaques is a tedious and expensive task that involves many stakeholders like the local public, professional monkey catchers and local authorities. Moreover, it is a multi-stage process that requires information of the target macaque troupes like their feeding location, number of individuals in the troupe, etc. With technology for face detection and recognition, we foresee a way of reducing the cost and effort in catching monkeys, especially in urban areas. 

In this paper, we presented MFID, a deep learning based automatic pipeline for macaque face recognition. We showed state of the art performances on a variety of experimental settings, including \emph{open set identification}, which is the most relevant in the context of identifying sterilized monkeys. Based on the robust performance of MFID, we plan to integrate it in a crowdsourced platform, like a mobile phone app. The app would allow humans to report monkey sightings and incidents in their neighborhood by uploading their images or videos to the cloud. MFID could then identify individuals and help estimate troupe strengths, as well as other meta-information like gender distribution, adult/pup ratios etc. The identified macaques in the troupes would be checked against a database of sterilized individuals and the appropriate authorities notified for further action. MFID could also be used by biologists for large-scale population monitoring surveys of macaques or other primate species. Keeping these implications in mind, we have chosen deep learning models that are relatively small in size and could potentially be run on a modern smartphone with moderate requirements. We have employed a DenseNet-121 network that takes only   $28.4$  MB of memory as opposed to other models like AlexNet, Resnet-50 and ResNet-18 that occupy $ 245 $ MB, $ 90 $ MB and $ 44 $ MB respectively. 

In the near future, we plan to integrate MFID with a crowdsourced mobile app, and use it for larger-scale data collection and evaluation. We hope to put this app to use by locals in the states of Uttarakhand and Himachal Pradesh in India for testing. These states report the largest number of incidents of human-monkey conflict. We anticipate strong participation from people in these states as they are the biggest stakeholders. Finally, we hope our current discussions with govt. authorities and biologists working with them will lead to adoption of MFID in the sterilization process. We plan to post the collected datasets in public forums and augment them as we collect more data from the field. We will also open source the code for the presented model of MFID.



\bibliographystyle{aaai}
\bibliography{references}
\end{document}